\pdfoutput=1
\documentclass[11pt,a4paper]{article}
\usepackage[hyperref]{emnlp2020}
\usepackage[utf8]{inputenc}
\usepackage[T1]{fontenc}
\usepackage{times}
\usepackage{latexsym}
\usepackage[flushmargin]{footmisc}

\usepackage{microtype}

\aclfinalcopy 
\def\aclpaperid{SIGTYP 5}

\title{Towards Induction of Structured Phoneme Inventories}

\author{Alexander Gutkin\\
  Google, UK\\
  {\small\texttt{agutkin@google.com}}\\\And
  Martin Jansche\thanks{\ \ This work was done while the author was at Google, prior to joining Amazon.}\\
  Amazon, UK\\
  {\small\texttt{jansche@amazon.com}}\\\And
  Lucy Skidmore\\
  University of Sheffield, UK\\
  {\small\texttt{lskidmore1@sheffield.ac.uk}}\\}

\date{}

\begin{document}
\maketitle

\vspace{-1.5mm}
\section{Introduction}
\vspace{-1.7mm}
Phonological typology is an important branch of linguistic typology
concerned with the study of the distribution and behavior of sounds in
world's languages~\cite{gordon2016}. The cross-linguistic typological
databases, such as \textsc{phoible} by~\citet{phoible}, provide the crucial tools
for drawing typological generalizations within the field. In addition
to facilitating development of probabilistic models of phonological
typology~\cite{cotterell-eisner-2017-probabilistic,ahn2019}, such
resources were also shown to positively influence the downstream
multilingual NLP~\cite{littell-etal-2017-uriel}, speech~\cite{li2020}
and language documentation tasks~\cite{anastasopoulos2019}.

This short paper provides an overview of our completed and ongoing
experiments with phonological representations utilizing phonological
typology databases in speech processing tasks. Despite the increasing
popularity of end-to-end approaches to automatic speech
recognition~\cite{moritz2019}, text-to-speech~\cite{chen2019} and
speech-to-speech translation~\cite{tjandra2019}, there are still
plenty of scenarios where integration of accurate phonological
knowledge is crucial, or at least beneficial, including the end-to-end
approaches themselves as recently demonstrated by~\citet{salesky2020}.

In particular, we describe two strains of research motivated by the
need to scale the development of speech technologies that still
require phonological representations to low-resource languages and
dialects. We first overview the framework for analysis of
cross-lingual consistency of phonological features in multilingual
phoneme inventories derived from cross-lingual typological
databases. We then offer a sketch of a method that may serve as a
potential building block in the future phoneme inventory induction
system and the central role the phonological typology plays in this
approach.

\vspace{-1mm}
\section{Multilingual Phoneme Inventories}
\vspace{-1.7mm}
The phoneme was originally defined as a theoretical abstraction that
applies language-internally. Using phonemes and their succinct
distinctive feature (DF) encodings in cross-linguistic settings, the
practice going back to~\citet{dalsgaard1992}, raises an important
question: given a multilingual phoneme inventory derived from a
typological database, such as \textsc{phoible} or
\textsc{PanPhon}~\cite{mortensen-etal-2016-panphon}, it is not clear a priori
whether all the DFs will be useful or even valid. If DF
representations were phonetic rather than phonemic, and acoustic
rather than articulatory, one would expect a close correspondence
between DFs and the acoustic signal. In practical multilingual
applications, however, the representations are often guided by purely
phonemic considerations because of the availability of phonemic
inventories and transcriptions. Such approaches are often outperformed
by the models striving for more phonetic realism, such as the
allophonic models of~\citet{mortensen2020}.

We followed a simple method: to consider a phonemic contrast to be
consistent or robust across languages, it needs to be easily predictable
on heldout languages in a binary classification
task~\cite{johny2019}. An instance of this problem consists of a span
of a speech signal (e.g., a vowel in surrounding context) and a
positive or negative label (e.g., front vowel vs. back vowel). A
classifier is trained on a multi-speaker, multi-language dataset
withholding one or more languages, which are later used for
evaluation. For cases where cross-linguistic consistency did not
hold, we extended this method by additionally grounding the
representation on the contextual phonological knowledge given as DFs,
excluding the contrast itself~\cite{skidmore2020}. For our experiments
we used a set of languages from Dravidian, Indo-European
and Malayo-Polynesian families with the phoneme inventories derived
from \textsc{phoible}.

Overall, our findings are mixed. A specially designed experiment for
predicting contrasts in unvoiced labial consonants between Bengali and
Spanish produced consistent and cross-lingually robust
predictions~\cite{johny2019}, also for a variety of auditory
representations~\cite{gutkin2020}, despite the conflicting status of
some of the allophones of the phonemes in the experiment. Similarly
robust were contrasts between front and back vowels, as well as the
vowel height and continuant manner of articulation
distinctions~\cite{skidmore2020}. The negative results include the
cross-lingual prediction of retroflex consonants between the language
families: retroflex predictor trained on Dravidian languages fails to
reliably predict retroflex consonants in Bengali, conversely the
predictor trained on Indo-Aryan languages is not reliable for
Malayalam. Similarly inconsistent, but less disappointing, was the
detection of aspiration. Furthermore, inclusion of other contrasts as
contextual input features did not lead to significant improvements in
predicting these hard cases.

One of the motivations behind our investigations describe above is a
research question that still remains open: Can the above methodology
be used for analyzing the cross-linguistic quality of the existing
phoneme inventories given the data? For example, among the
Malayo-Polynesian languages we considered, only Javanese has
retroflex consonants, which it acquired through loanwords from
Indo-Aryan or Dravidian languages~\cite{ogloblin2006}. \textsc{phoible}
contains three different phoneme inventories for Javanese, the largest
of which \texttt{GM} \texttt{1675} represents retroflex plosives,
while the smallest \texttt{UPSID} \texttt{380}~\cite{phoible380} omits
them. Which of the two representations is more likely to be a better
fit in a multilingual pronunciation model?

\vspace{-1mm}
\section{Towards Phonology Induction}
\vspace{-1.7mm}
Our ongoing research focuses on induction of phoneme inventories for
languages for which the conventional resources required for speech
model training are often missing, placing this task among other
related work in zero-resource subword modeling
problems~\cite{baljekar2015,lee2015,chen2017}. The current
state-of-the-art unsupervised acoustic unit discovery approaches
derive acoustic-phonetic~\cite{hu2020,morita2020,niekerk2020} and
latent auditory-like~\cite{ondel2019} representations, yet it is
unclear how accurate these representations are from a typological
standpoint.

Our initial investigations utilized ``universal'' multilingual phoneme
recognizers. The results were disappointing primarily because the
training data sparsity and the absence of language models to constrain
the search often resulted even in unreliable recovery of the phoneme
inventories for unseen dialects of the same language, e.g.,
determining the inventory of Argentinian Spanish having seen Castilian
and Mexican Spanish. A more viable approach to this task is
integrating language identification and phonological typology into the
phoneme recognizer. Given an accurate language identification model,
the \textsc{phoible} phoneme inventories belonging to the ``closest''
languages may be used to constrain the phonemic hypothesis space for
the unseen language or dialect. We developed efficient search
techniques for the lattices of this type~\cite{jansche2019}.

Currently we are revisiting the phonological contrast predictor
methods described in the previous section to adapt them to phonology
induction tasks. Several known approaches to phonemic feature
detection in continuous speech are known, some are purely reliant on
signal processing, while others are
model-based~\cite{frankel2007,muller2017}. At the simplest level, the
output of such predictors represents speech as the parallel
asynchronous streams of articulatory features. More sophisticated
models that exploit the structure of articulation, feature geometry
and other types of correlations between various features are also
possible. In these approaches, the phonological cross-linguistic
databases, such as \textsc{phoible} or \textsc{PanPhon}, provide the important source
of truth not only for combining various features into phonemes, but
also for determining which combinations of hypothesized phonemes are
admissible given the existing phoneme inventories. Furthermore,
additional phonological evidence provided by other typological
resources, such as World Atlas of Language Structures
(\textsc{WALS})~\cite{wals}, can be integrated as well if it can be
reliably extracted from the speech signal~\cite{gutkin2018}.

\bibliographystyle{acl_natbib}
\bibliography{anthology,main}

\end{document}